\documentclass[10pt,twocolumn,letterpaper]{article}

\usepackage{cvpr}              

\usepackage{pifont}
\newcommand{\cmark}{\ding{51}}
\newcommand{\xmark}{\ding{55}}

\definecolor{cvprblue}{rgb}{0.21,0.49,0.74}
\usepackage[pagebackref,breaklinks,colorlinks,allcolors=cvprblue]{hyperref}

\def\paperID{10113} 
\def\confName{CVPR}
\def\confYear{2026}

\title{Learning a Particle Dynamics Model with Real-world Videos}

\author{
\begin{tabular}{c}
Chanho Kim \quad Suhas V. Sumukh \quad Li Fuxin \\
[0.2em]
Oregon State University \\
{\tt\small \{kimchanh, suhas.sumukh, fuxin.li\}@oregonstate.edu}
\end{tabular}
}

\begin{document}
\maketitle
\begin{abstract}
Data-driven learning approaches for physics simulation, sometimes referred to as world models, have emerged as promising alternatives to traditional physics simulators due to their differentiable nature. Prior work has demonstrated impressive results in predicting the motions of rigid and non-rigid objects in complex scenes involving multiple interacting bodies. However, these models are typically trained in simulated environments because obtaining perfect state information such as complete scene point clouds and point correspondences over time is challenging in real-world settings. This reliance on synthetic data can limit their applicability when the sim-to-real gap is large. In this work, we aim to overcome these limitations by introducing a novel framework for training neural object dynamics models directly from unlabeled real-world videos. Specifically, we propose to learn a particle-based dynamics model compatible with a Gaussian splatting framework, which operates on dense particles derived from Gaussians (i.e., particles with scales and rotations) and predicts their position and rotation changes over time. The model is trained via rendering supervision, enabling learning from real-world videos without requiring particle-level labeled states. Our model operates directly on dense Gaussians without relying on heuristic subsampling anchor points. To enable this study, we also present a real-world dataset consisting of about 500 videos capturing diverse object interactions. The dataset and code will be publicly released at our project webpage: \href{https://chkim403.github.io/gs_physics}{\texttt{https://chkim403.github.io/gs\_physics}}.
\end{abstract}    
\section{Introduction}
\label{sec:intro}

\begin{figure*}[t]
        \centering
        \includegraphics[width=1.0\textwidth]{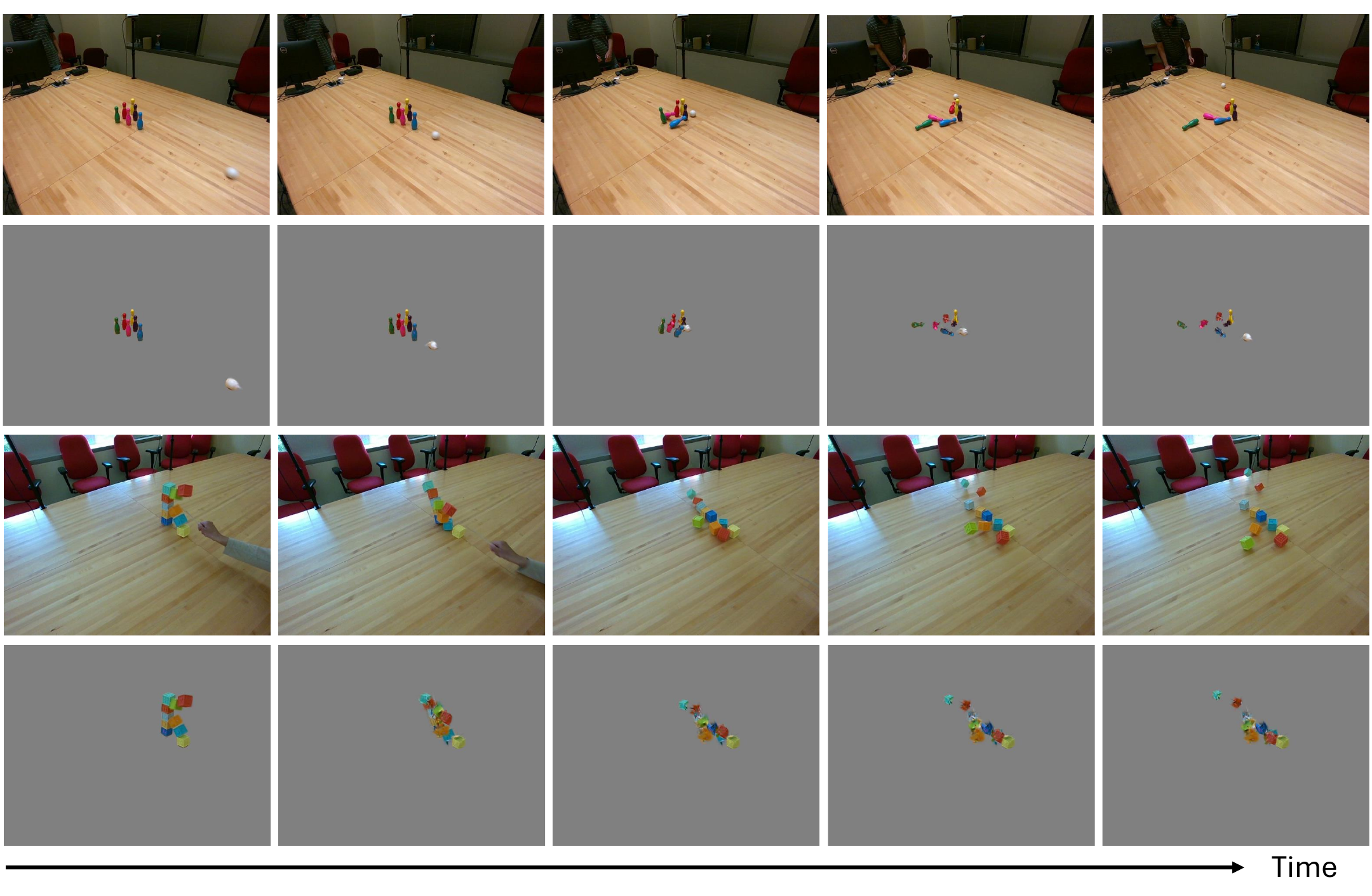}
        \caption{Example sequences illustrating the physical scenarios of interest. The dataset captures multi-object interactions with complex collision dynamics, including bowling-style impacts and cube-stack collapses. The first and third rows show the original RGB sequences, while the second and fourth rows show renderings from the 3D Gaussian rollouts predicted by our dynamics model for the same viewpoints. Our approach enables training dynamics models in multi-object settings directly from real-world videos.}
        \label{fig:intro}
    \end{figure*}

Humans develop an intuitive understanding of physics early in life simply by observing the world, long before studying the physics subject in any school. This innate ability enables us to effortlessly track the motion of nearby objects and make reasonably accurate predictions about their future trajectories based on past movements. Such skills are essential for acting safely and effectively in a complex environment where constant interactions with other objects and people are inevitable. 

Endowing AI agents with similar physical reasoning capabilities could benefit numerous applications such as manufacturing, robotics, and generative modeling \cite{allen2022physical, zhang2024particle, huang2025particleformer, nvidia2025cosmosworldfoundationmodel}. One approach to achieve this is through classical particle-based simulations, such as the Material Point Method (MPM) ~\cite{hu2019chainqueen, Xie_2024_CVPR}. However, these methods typically require all physical parameters of the particles to be predefined, an assumption that rarely holds in real-world scenarios. 

Alternatively, learning particle dynamics directly using neural networks has attracted considerable attention due to its efficiency and differentiability, which makes such models more suitable as components of larger end-to-end trainable systems. Despite this promise, existing models remain limited in several ways. Many methods \cite{kim2024object, han2022learning, pfaff2021learning, sanchez2020learning, li2019learning} are trained primarily on simulated data, as obtaining real-world 3D labeled data, such as dense point cloud correspondences across frames, is expensive and challenging, particularly for videos. Even when trained on real-world sequences \cite{shi2023robocook}, the supervision signals are often noisy because the loss relies on approximate distance functions such as the Chamfer distance. This lack of reliable supervision limits the applicability of current models in diverse real-world settings.

Recent advances in differentiable rendering such as Gaussian Splatting (GS) and Neural Radiance Fields (NeRF) \cite{kerbl3Dgaussians, mildenhall2020nerf} present a promising alternative. By enabling gradients to flow from 2D observations back to 3D, these methods allow 3D models to be \textbf{supervised directly from images}. This opens the door to learning dynamics from real videos without ground-truth 3D labels or a physics simulator, where there are \textbf{much more} data available online to scale training. Yet prior work in this direction has focused almost exclusively on single-object scenes \cite{lu2025gaussianworldmodel, zhang2024particle, zhang2024dynamic, 22-driess-NeRF-RL}, hence does not capture the complex discontinuous interactions required for multi-object collision dynamics.

In this work, we take a step on learning \textbf{multi-object collision dynamics} directly from \textbf{real world videos} via rendering-based supervision, a capability that was previously unavailable to the community to the best of our knowledge. We propose a model that learns the collision dynamics of multiple interacting objects using only 2D object masks from a SAM-style video segmentation model \cite{ravi2024sam2, dam4sam}, combined with a differentiable rendering loss from multiple calibrated viewpoints. Learning in this setting introduces several challenges absent in prior work: 3D trajectory extraction from partial 2D cues, assigning Gaussians to different objects through occlusions and collisions, and predicting future Gaussian states in a feed-forward manner.

To support this problem, we also collect a new real-world multi-view dataset (Fig.~\ref{fig:intro}) containing about 500 videos captured from four calibrated cameras across two challenging scenarios—falling cube stacks and bowling—that exhibit rich, nontrivial multi-object interactions rarely available in prior datasets.

In summary, our contributions are:
\begin{itemize}
\item We presented the first approach to extend multi-object, action-free 3D particle dynamics models to training on real videos with rendering loss without knowledge of the physical properties of the particles.
\item We capture around 500 multi-view videos with complex contact events that serve as a benchmark for learning real-world physical interactions.
\item We provide  reference designs and ablations for key system components, including Gaussian trajectory initialization, assigning Gaussians to objects and multi-step rollout training. Code and dataset will be released.

\end{itemize}
\section{Related Work}
\label{sec:related_work}

Recent advances in rendering-based 3D reconstruction such as Gaussian Splatting (GS) and Neural Radiance Fields (NeRF) \cite{kerbl3Dgaussians, mildenhall2020nerf} have opened new possibilities for leveraging 2D image observations as ground-truth supervision signals for 3D models. Several prior works have leveraged NeRF to train GNN-based dynamics models via rendering supervision \cite{22-driess-NeRF-RL, xue2023dintphys, whitney2024modeling}. However, NeRF is inherently a volumetric model and can only model particle dynamics through an implicit deformation field, which can be more ill-posed than necessary and it is difficult to add a notion of object among particles or impose rigidity constraints.

Gaussian Splatting (GS) \cite{kerbl3Dgaussians} offers a more explicit scene representation, making it possible to use well-established architectures such as GNNs and point-cloud backbones to process 3D Gaussians directly. 
Recent work in robotics \cite{lu2025gaussianworldmodel, zhang2024particle, zhang2024dynamic} has explored training world models directly from real world videos, with the GS framework either enabling rendering-based supervision \cite{lu2025gaussianworldmodel, zhang2024dynamic} or enabling test-time video generation \cite{zhang2024particle}. However, these approaches primarily model the effects of robot actions on a single manipulated object, limiting their applicability to specific task contexts.

For multi-object collision scenarios, rendering-based supervision has also been used to learn dynamics models for complex collisions across multiple interacting objects \cite{allen23a, NEURIPS2024_510cfd99, whitney2024learning, zhobro2025learning3dgaussiansimulatorsrgb}, but prior work has been limited mostly to simulated data, where rendering-based loss makes less sense due to the availability of better ground truth. \cite{allen23a} demonstrated learning collision dynamics from real world labeled videos, but their dataset contained only simple scenes involving a single cube being thrown onto a table, and the ground truth cube poses were obtained using visual markers placed on the cube. To our knowledge, our work is the first to learn non-trivial multi-object collision dynamics directly from real-world videos without markers. 

Gaussian Splatting has also been used to generate simulatable assets, where Gaussians extracted from real-world observations are augmented with object properties. These properties are either user-specified \cite{Xie_2024_CVPR} or estimated from video observations \cite{jiang2025phystwin, zhong2024springgaus}. The resulting simulatable Gaussians can then be used to produce rollouts under arbitrary forces through MPM or Euler integration. In contrast to these approaches, we aim to learn a dynamics model that generates rollouts directly, without any manual tuning or additional system identification steps at test time.
\section{Method}

We base our approach on prior work~\cite{kim2024object} that utilizes a point cloud backbone \cite{wu2019pointconv,wu2023pointconvformer} for object interaction modeling with point cloud input. A set of 3D Gaussians produced by Gaussian Splatting is essentially a 3D point cloud where each point is associated with additional attributes such as scale and rotation. Therefore, in this work we directly treat Gaussians as points and apply a point cloud backbone. 

In Sec. 3.1--3.2, we introduce our problem setup. In Sec. 3.3, we briefly review the prior work that forms the basis of our convolution-based interaction modeling. In Sec. 3.4--3.6, we present our approach for enabling GS-based collision dynamics learning.

\subsection{Gaussian Representation}

We start with a set of Gaussians that are produced by an external Gaussian Splatting algorithm \cite{kerbl3Dgaussians} on a single frame.
Each Gaussian $i$ is associated with its center $\mathbf{x}^{(i)}\in\mathbb{R}^3$, rotation $\mathbf{R}^{(i)}\in SO(3)$, scale $\mathbf{s}^{(i)}\in\mathbb{R}^3$, color $\mathbf{c}^{(i)}\in\mathbb{R}^3$, and opacity $o^{(i)}\in\mathbb{R}$:
\begin{equation}
G^{(i)} = \big(\mathbf{x}^{(i)},\ \mathbf{R}^{(i)},\ \mathbf{s}^{(i)},\ \mathbf{c}^{(i)},o^{(i)}\big).
\end{equation}
Gaussian Splatting allows these Gaussians to be rendered into images via a differentiable Gaussian renderer, which enables rendering-based supervision.

The rendering equation for a pixel $I$ is:
\begin{equation}
I(u) = \sum_{j =1}^J \mathbf{c}^{(j)} \alpha_c^{(j)}(u) \prod_{k=1}^{j-1}(1-\alpha_c^{(k)}(u))
\end{equation}
where all the $J$ Gaussians are sorted by the distance from their centers to the camera, and $\alpha_c^{j}(u)$ is the opacity $o^{(j)}$ multiplied with the projection of the Gaussian $G^{(j)}$ density along the camera ray that passes through $u$.

Let the three input frames be at times $t-2, t-1, t$. For each Gaussian $i$ in frame $t$, we augment its representation with its two most recent velocities
\begin{equation}
\mathbf{v}^{(i)}_{t-1} = \mathbf{x}^{(i)}_{t-1}-\mathbf{x}^{(i)}_{t-2}, \qquad
\mathbf{v}^{(i)}_{t}   = \mathbf{x}^{(i)}_{t}-\mathbf{x}^{(i)}_{t-1}.
\end{equation}
which can be obtained from any video via point tracking approaches such as~\cite{harley2025alltracker} (see Sec.~\ref{sec:data_collection} for more details). We choose this representation which does not require space-time Gaussians because we were not able to obtain reliable space-time Gaussians with available methods~\cite{luiten2023dynamic,zhang2024dynamic}.

\subsection{Problem Formulation}
Besides the two most recent velocities, we also extract the two most recent vertical coordinates $z^{(i)}_{t-1}$ and $z^{(i)}_{t}$ from each Gaussian (which are informative for modeling gravity and ground plane). We concatenate these into the per--point feature vector
\begin{equation}
\mathbf{f}^{(i)}_t = \big[\,\mathbf{v}^{(i)}_{t-1},\ \mathbf{v}^{(i)}_{t},\ z^{(i)}_{t-1},\ z^{(i)}_{t}\,\big].
\end{equation}
Given the set of per-point features $\{\mathbf{f}^{(i)}_t\}_{i=1}^N$, the network predicts the next frame Gaussian center $\{\mathbf{x}^{(i)}_{t+1}\}$ and rotation $\{\mathbf{R}^{(i)}_{t+1}\}$.
The model can parameterize the center prediction either by predicting the next velocity or by predicting acceleration following \cite{allen23a}:
\begin{equation}
\hat{\mathbf{x}}^{(i)}_{t+1} = \mathbf{x}^{(i)}_t + \hat{\mathbf{v}}^{(i)}_{t+1}, \qquad \hat{\mathbf{x}}^{(i)}_{t+1} = \mathbf{x}^{(i)}_t + \mathbf{v}^{(i)}_t + \hat{\mathbf{a}}^{(i)}_t.
\end{equation}
The rotation is updated via
\begin{equation}
\mathbf{R}^{(i)}_{t+1} = \Delta\mathbf{R}^{(i)}_t \,\mathbf{R}^{(i)}_t.
\end{equation}
where $\Delta\mathbf{R}^{(i)}_t\in SO(3)$ is the predicted delta rotation.

\subsection{Convolution-based Interaction Modeling}
Prior work~\cite{kim2024object} modeled interactions among 3D points using PointConv~\cite{wu2019pointconv, wu2023pointconvformer}, which offers an efficient convolution operation over continuous 3D point sets. The 3D convolution operation is written as:
\begin{equation}
\mathbf{y}_0 = \mathbf{W}_l \mathrm{vec}\left(\frac{1}{|\mathcal{N}(\mathbf{p}_0)|} \sum_{\mathbf{p}_i \in \mathcal{N}(\mathbf{p}_0)} h(\mathbf{p}_i - \mathbf{p}_0) \mathbf{x}_i^\top\right)
\label{eq:pointconv}
\end{equation}
where $\mathbf{p}_0\in \mathbb{R}^3$ is a query point location for output features $\mathbf{y}_0 \in \mathbb{R}^{c_{\text{out}}}$. For each neighboring point $\mathbf{p}_i \in \mathbb{R}^3$ in the neighbor set $\mathcal{N}(\mathbf{p}_0)$, $\mathbf{x}_i \in \mathbb{R}^{c_{\text{in}}}$ denotes the input feature associated with that point, and $|\mathcal{N}(\mathbf{p}_0)|$ is the neighborhood size. $\mathbf{W}_l \in \mathbb{R}^{c_{\text{out}} \times (c_{\text{in}} c_{\text{mid}})}$ is a shared linear layer, $\mathrm{vec}(\cdot)$ converts a matrix into a vector, and $h(\cdot) \in \mathbb{R}^{c_{\text{mid}}}$ is an MLP that generates convolution kernel weights by embedding the relative position $(\mathbf{p}_i-\mathbf{p}_0)$.

The network in~\cite{kim2024object} interleaves Object PointConv and Relational PointConv layers. Object PointConv applies 3D continuous point convolution over points belonging to the same object, while Relational PointConv applies convolution between a point from one object and nearby points from other objects. Thus, in Eq.~(\ref{eq:pointconv}), the neighbor set $\mathcal{N}(\mathbf{p}_0)$ alternates between these two types depending on the layer, enabling the model to effectively learn force propagation effects both within objects and across objects.

\subsection{Gaussian Integration}

Unlike the perfect particle-level ground truth where each point is associated with a correct object ID in simulation environments, Gaussians produced by GS do not come with object IDs. However, we assume having  multi-view 2D object segmentation masks that are consistent across camera views and over time, which are much more realistic thanks to foundational 2D segmentation models~\cite{ravi2024sam2,dam4sam} (data collection details in Sec.~\ref{sec:data_collection}). We therefore extract object identity information of each Gaussian by measuring how much each Gaussian contributes to rendering each object mask.

In Gaussian Splatting, each Gaussian contributes to the color of multiple pixels. The rendering contribution of Gaussian $i$ to pixel $u$ in view $c$ is written as:
\begin{equation}
\gamma^{(i)}_{c}(u)
=
\alpha^{(i)}_{c}(u)\;
\prod_{j \in \mathcal{F}_{c}(i)} \bigl(1 - \alpha^{(j)}_{c}(u)\bigr),
\end{equation}
where $\mathcal{F}_{c}(i)$ denotes the set of Gaussians that are in front of Gaussian $i$ in view $c$.

For each object mask $m$ in view $c$, we take the maximum rendering contribution of Gaussian $i$ over all pixels $u$ inside that mask:
\begin{equation}
\Gamma^{(i)}_{c,m} \;=\; \max_{u \in \text{mask}(c,m)} \gamma^{(i)}_{c}(u).
\end{equation}
Given $N$ Gaussians, $C$ views, and $M$ object masks per view, we obtain a tensor $\Gamma \in \mathbb{R}^{N \times C \times M}$ where each entry $\Gamma^{(i)}_{c,m}$ quantifies the strongest rendering evidence that Gaussian $i$ belongs to object $m$ in view $c$.

To assign an object ID to each Gaussian, we aggregate $\Gamma$ across all camera views using a majority voting scheme. For each Gaussian $i$ and each view $c$, we determine the most likely object ID for that view by
\begin{equation}
\text{ID}^{(i)}_c \;=\; \arg\max_{m \in \{1,\dots,M\}} \Gamma^{(i)}_{c,m}.
\end{equation}
We aggregate votes to determine the most frequently occurring ID across views:
\begin{equation}
\mathbf{v}^{(i)}[m] \;=\; \sum_{c=1}^{C} \mathbf{1}\!\left\{ \text{ID}^{(i)}_c = m \right\}
\end{equation}
where $\mathbf{v}^{(i)} \in \mathbb{R}^M$ is the vote count vector across views, and $\mathbf{1}\{\cdot\}$ denotes the indicator function, which is $1$ if the condition is true and $0$ otherwise. We convert the vote counts to a one--hot vector by
\begin{equation}
\mathbf{w}^{(i)} = \text{one-hot}\!\left(\arg\max_m \mathbf{v}^{(i)}[m]\right).
\label{eq:discrete_id}
\end{equation}
Alternatively, one can obtain a soft distribution instead of a one--hot vector. We provide ablations comparing different ID extraction approaches in the experiments section.

Unlike~\cite{kim2024object}, where explicit object-specific neighborhoods were maintained to define nearest neighbors from the same object or different objects, we perform an object-agnostic KNN search first, and use the ID vectors to decide which neighbors contribute to each convolution. For Object PointConv, we define the affinity weight $m_{i,0} = (\mathbf{w}^{(i)})^\top \mathbf{w}^{(0)}$, and for Relational PointConv, we define $m_{i,0} = 1 - (\mathbf{w}^{(i)})^\top \mathbf{w}^{(0)}$. We then plug $m_{i,0}$ into Eq.~(\ref{eq:pointconv}) as:
\begin{equation}
\mathbf{y}_0 = \mathbf{W}_l \,\mathrm{vec}\!\left(\frac{1}{\sum_{i} m_{i,0}} \sum_{\mathbf{p}_i \in \mathcal{N}(\mathbf{p}_0)} m_{i,0}\, h(\mathbf{p}_i - \mathbf{p}_0)\, \mathbf{x}_i^\top \right).
\label{eq:pointconv_attn}
\end{equation}
This enables us to form the same object-level partitions that the convolution layers in Sec.~3.3 require (Object PointConv vs.\ Relational PointConv), without explicitly computing object-level neighborhoods, which would otherwise require inefficient per-object looping. This approach also naturally supports both discrete one-hot vectors and soft distributions for representing each Gaussian's object ID.

\subsection{Network Architecture}

We use a U-Net architecture that takes the original dense Gaussians as input and constructs hierarchical scene representations via two grid downsampling layers with grid sizes of $2$\,cm and $5$\,cm, while interleaving Object and Relational PointConv layers throughout the hierarchy, similar to~\cite{kim2024object}. At the final resolution, a prediction head operating at the original dense Gaussian locations predicts the next velocity. The rotation update is predicted either by a follow-up rigid transformation fitting module \cite{umeyama_91} if it is enabled or directly by the prediction head using the 6D rotation representation~\cite{Zhou_2019_CVPR}. Additional architectural details are provided in the supplementary material.

\subsection{Training Details} \label{sec:training}

We train our model using two types of supervision: rendering supervision and position regression. Neither source provides perfect or noise-free guidance. For rendering supervision, although we have clean image observations as ground truth, the extracted 3D Gaussians are imperfect, making it impossible to render future frames perfectly even with accurate motion predictions. For position supervision, the pseudo-labels obtained from long-term point tracking and video object segmentation also contain inherent noise. Therefore, we rely on both sources of supervision to train our model.

From the three input frames, we predict $K$ future frames using our model. In this work, we set $K=3$ for all experiments. During testing, the rollout can be extended for as long as needed. For $K>1$, the prediction from the network is fed back into the model as input in an autoregressive manner. Given predictions for $K$ frames of $B$ scenes in a mini-batch, our loss function is written as:
\begin{equation}
\mathcal{L} = \frac{1}{BK} \sum_{i=1}^{B}\sum_{k=1}^{K} \left(
\lambda_{\text{rend}}\, \mathcal{L}^{(i,k)}_{\text{render}}
\;+\;
\lambda_{\text{pos}}\, \mathcal{L}^{(i,k)}_{\text{pos}}
\right).
\end{equation}
Here, $\mathcal{L}^{(i,k)}_{\text{render}}$ is the rendering loss for the $k$-step prediction of the $i$-th example, and $\mathcal{L}^{(i,k)}_{\text{pos}}$ is the position regression loss for the $k$-step prediction of the $i$-th example. We use an L1 loss for $\mathcal{L}_{\text{render}}$ and the Huber loss for $\mathcal{L}_{\text{pos}}$. We set $\lambda_{\text{rend}} = 3$ and $\lambda_{\text{pos}} = 1$.

For training, we define one epoch as the iteration in which each scene is sampled 10 times when forming mini-batches. Frames are sampled either uniformly at random or according to a loss-based distribution which prioritizes higher-loss frames (i.e., hard example mining). All models reported in Sec.~\ref{sec:experiment} are trained for 50 epochs. We use a batch size of $B=12$ and train all models for 50 epochs with an initial learning rate of $0.001$ using the Adam optimizer~\cite{kingma2015adam}. A step-based learning rate scheduler is applied, reducing the learning rate by a factor of 0.1 twice during training.

\section{Real-World Collision Dataset}

\begin{figure}[t]
        \centering
        \includegraphics[width=0.47\textwidth]{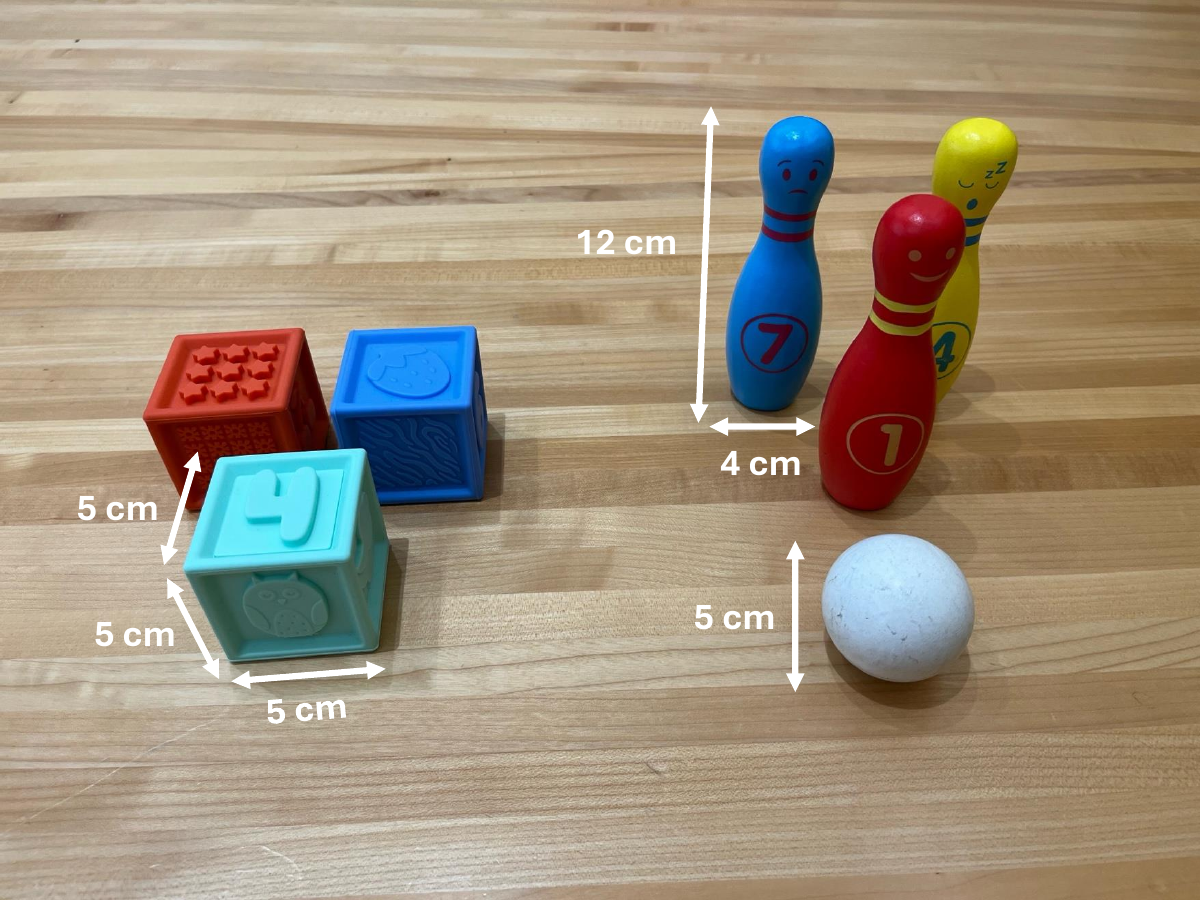}
        \caption{Objects used in our dataset. The falling-cube-stack scenario includes up to 10 cubes, while the bowling scenario includes one ball and up to 10 bowling pins. Overlaid annotations indicate real-world object dimensions.}
        \label{fig:toy_set}
    \end{figure}

We present the first real-world dataset for learning multi-object collision dynamics, consisting of two distinct scenarios: falling cube stacks and bowling. In the falling cube stack scenario, towers built from toy cubes are subjected to a randomly applied external force, causing them to collapse. In the bowling scenario, a ball impacts up to ten toy bowling pins. Both scenarios are recorded on a tabletop setup using the toy sets shown in Fig. \ref{fig:toy_set}. With this, we hope to build a robust data collection pipeline that can easily scale up to more scenarios. We introduce more details about the dataset and how we label our data to enable model training. 

\subsection{Dataset Details}

\begin{figure}[t]
        \centering
        \includegraphics[width=0.48\textwidth]{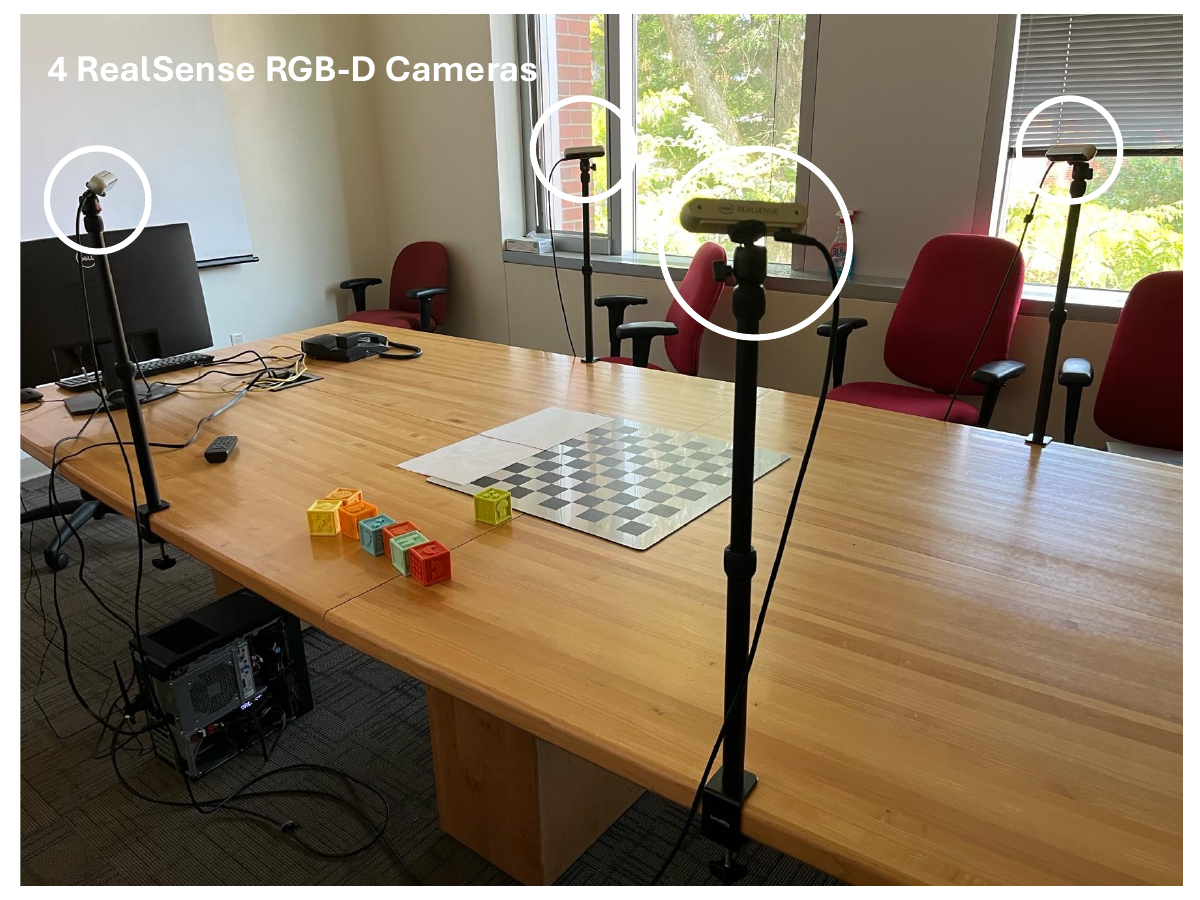}
        \caption{Data collection setup with four Intel D455 RealSense cameras. Camera intrinsics and extrinsics are estimated prior to recording using a checkerboard calibration.}
        \label{fig:setup}
    \end{figure}

We collected $210$ falling cube stack sequences and $292$ bowling sequences. All sequences were captured in a calibrated multi-view setting using four Intel RealSense D455 cameras arranged around the table (Fig.~\ref{fig:setup}). Camera intrinsics and extrinsics were estimated using a checkerboard placed on the table prior to recording. Images were recorded at a resolution of 640×480. Each scene consists of synchronized RGB-D images with known camera poses, as well as per-frame object segmentation masks whose object IDs are consistent across all camera views and over time. Additional details on data cleaning, annotation, and mask generation are provided in the next subsection.

\subsection{Data Collection Pipeline}
\label{sec:data_collection}

\begin{figure*}[t]
        \centering
        \includegraphics[width=1.0\textwidth]{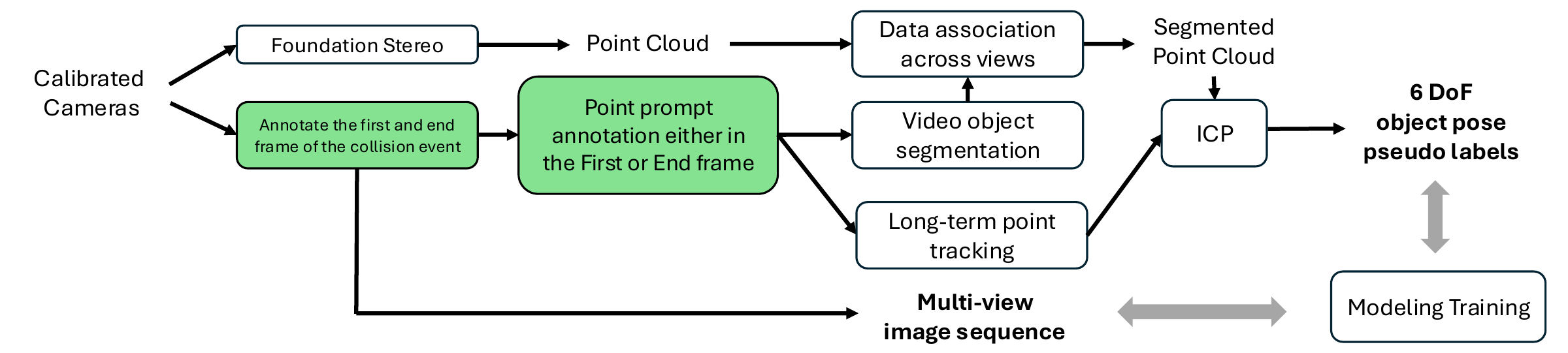}
        \caption{An overview of our data collection pipeline. It enables learning collision dynamics from real-world videos by providing two complementary learning signals for the model. Steps requiring manual annotation are highlighted in green.}
        \label{fig:pipeline}
    \end{figure*}

The overall data processing pipeline is shown in Fig. \ref{fig:pipeline}. After recording, each scene contains RGB-D image sequences from four calibrated cameras. For depth, instead of using the noisy depth maps produced by the RealSense cameras, we generate depth maps using FoundationStereo \cite{wen2025stereo} from the RealSense left and right IR cameras. During capture, we disable the RealSense IR emitter, since the projected dot patterns may degrade FoundationStereo performance and limit its ability to generalize to our scenes.

For each recording, we manually annotate the first and last frame such that each sequence only contains the portion of the interaction after the initial impulse is applied by the human operator. In addition, we manually annotate a small set of 2D points (per object) at either the first or the last frame depending on visibility, and use these points as prompts for a SAM-style video segmentation and tracking method \cite{dam4sam, ravi2024sam2}. Based on the annotation frame, masks are propagated either forward or backward through the sequence. To ensure consistent object IDs across all four camera views within a scene, we perform an additional cross-view data association step. Specifically, we select a reference view and find correspondences between object masks in the reference view and all other views via Hungarian matching, using the averaged 3D world coordinates of the 2D masks (e.g., 3D coordinates of pixels inside each 2D mask are obtained via depth and camera intrinsics/extrinsics and averaged). This produces multi-view object segmentation masks with consistent IDs semi-automatically, requiring only small amounts of manual annotations. While failures may occur (e.g., segmentation drift during tracking or inaccurate initial segmentation produced by point prompts), in practice we find such errors to be infrequent and acceptable as noise in the training data. 

Our network takes 3D Gaussian trajectories as input. We initially experimented with 4D Gaussian Splatting methods designed for dynamic scenes \cite{luiten2023dynamic, Wu_2024_CVPR}. However, we found these methods to be slow, and the resulting 3D Gaussian trajectories to be unreliable. They often look reasonably only rendered in 2D, but the underlying 3D motion is inaccurate. Therefore, instead of extracting dynamic Gaussians directly, we estimate per-object 3D pose changes over time and apply those transforms to static 3D Gaussians to obtain the Gaussian trajectories.  

We estimate per-object 3D poses using the iterative closest point (ICP) algorithm on partial point clouds in the world coordinate frame. ICP alone can fail when the initial alignment between two point clouds is poor. To obtain better initialization, we run an off-the-shelf 2D point tracking model \cite{harley2025alltracker} to produce point-level correspondences over time. Because each 2D pixel is associated with a 3D coordinate (via depth and camera calibration), we can lift tracked pixels into the 3D world frame, which gives us reliable initial correspondences. We initialize ICP with these correspondences and run ICP to convergence to recover the per-object pose trajectory. Once per-object poses are estimated for all frames, we propagate the static 3D Gaussians from any selected input frame through time to generate the 3D Gaussians trajectories used by our network. We generate the static 3D Gaussians in all frames by running Gaussian Splatting \cite{kerbl3Dgaussians} on each scene, and store them as part of the dataset. This allows us to generate training examples from a random frame during training.

\section{Experiment} \label{sec:experiment}

In this section, we detail our experimental settings and report performance comparisons on our dataset, along with comprehensive ablation studies. We conduct experiments on the real-world dataset described in Sec.~\ref{sec:data_collection}. For every video frame, we extract a set of 3D Gaussians, ensuring that the model always receives Gaussian-based inputs during both training and testing. Unless otherwise noted, all ablation baselines are trained under identical settings, and differences arise only from the variables we modify for each study. We run each method with three random seeds and report the mean and standard deviation.

\subsection{Performance Metrics}
Since we do not have ground-truth labels for evaluating the 3D reconstruction quality of each rollout, we primarily rely on rendering-based metrics such as PSNR, SSIM~\cite{1284395}, and LPIPS~\cite{zhang2018perceptual}, following prior work in 4D scene reconstruction~\cite{zhang2024dynamic, Wu_2024_CVPR}. In addition, because we have pseudo ground-truth particle-motion labels generated from long-term point tracking, video segmentation, and stereo-depth estimation, we also evaluate rollouts using the position-accuracy metric $\delta_{avg}$, which is commonly used in long-term point tracking~\cite{harley2025alltracker, luiten2023dynamic}. Due to the noise in the pseudo labels, we compute position accuracy at relatively high thresholds of $5$\,cm, $10$\,cm and $20$\,cm and report the average across the thresholds. Finally, we compute the Chamfer Distance (CD) (in cm) between the Gaussian positions produced by the rollout and those obtained during data preprocessing via GS. The CD is computed on the last frame of each sequence for which precomputed Gaussians are available, where these Gaussians serve as pseudo ground truth.

\subsection{Ablation Study}

We investigate several design choices to better understand their impact on final performance and report the results in Table~\ref{table:ablation}. From the overall results, we first note that rendering metrics are not very informative for evaluating neural dynamics models in our ablation setup, where the same input Gaussians are used across different methods. In contrast, CD and $\delta_{avg}$ are more discriminative, although these metrics are also imperfect since they rely on the pseudo ground truth. We discuss the limitations of current performance metrics in more detail in Sec.~\ref{sec:limitations}.

For the bowling scenario, the non-object-centric version of our network (i.e., without object–relation decomposition and without object-wise pose fitting) leads to unstable training and worse performance in terms of CD and $\delta_{avg}$, as shown in the first row of the table. Similarly, removing object-wise pose fitting alone negatively impacts these metrics, as shown in the second row.

Next, since object IDs are assigned based on the rendering contribution of each Gaussian to segmentation masks across multiple views, we compare two alternatives: using a discrete object ID obtained via majority voting, and using a soft object-distribution representation (both supported by our model as described in Eq.~(\ref{eq:pointconv_attn})). This comparison corresponds to the third row and the second-to-last row of the table. Overall, the model using discrete IDs performs better than the one using soft IDs. This is likely because a single Gaussian typically represents only one moving object, except in rare boundary cases.

Regarding the loss design, using both position and rendering losses leads to more balanced performance across both the bowling and cube stacks scenarios, as shown in the second-to-last row. Since each scene contains multiple frames and only a subset is sampled per epoch, we further compare two frame selection strategies: uniform sampling and hard example mining. For hard example mining, after each epoch we evaluate the model on all frames and construct a sampling distribution proportional to per-frame loss, so that more difficult frames are sampled more frequently. As shown by comparing the second-to-last and last rows of Table~\ref{table:ablation}, hard example mining improves performance in terms of CD and $\delta_{avg}$ but is not statistically significant.

\begin{table*}[t]
\centering
\caption{Ablation results. The “Components” column includes the following abbreviations: Pos. Loss (position loss), Rend. Loss (rendering loss), O-centric (object-centric modeling), Disc. ID (discrete object ID), P. Fit (pose fitting), and HEM (hard example mining). “--” denotes components that are not applicable under a given configuration. Results are bolded if they are significantly better via an unpaired t-test.} \label{table:ablation}
\scalebox{0.57}{
\setlength{\tabcolsep}{3pt}
\begin{tabular}{cccccc ccccc ccccc}
\toprule
\multicolumn{6}{c}{Components} & \multicolumn{5}{c}{Bowling} & \multicolumn{5}{c}{Falling Cube Stacks} \\
\cmidrule(lr){1-6} \cmidrule(lr){7-11} \cmidrule(lr){12-16}
Pos.\ Loss & Ren.\ Loss & O-Centric & Disc. ID & P. Fit & HEM 
& PSNR $\uparrow$ & SSIM $\uparrow$ & LPIPS $\downarrow$ & CD $\downarrow$ & $\delta_{avg}$ $\uparrow$
& PSNR $\uparrow$ & SSIM $\uparrow$ & LPIPS $\downarrow$ & CD $\downarrow$ & $\delta_{avg}$ $\uparrow$ \\
\midrule

\cmark & \cmark & \xmark & -- & \xmark & \xmark
& $\mathbf{27.39} \pm 0.08$ & $0.975 \pm 0.001$ & $0.053 \pm 0.004$ & $17.06 \pm 11.02$ & $56.74 \pm 8.12$
& $\mathbf{26.14} \pm 0.08$ & $0.973 \pm 0.000$ & $0.056 \pm 0.000$ & $10.76 \pm 0.68$ & $59.58 \pm 0.93$ \\

\cmark & \cmark & \cmark & \cmark & \xmark & \xmark
& $\mathbf{27.41} \pm 0.10$ & $0.976 \pm 0.001$ & $0.052 \pm 0.003$ & $11.15 \pm 2.04$ & $55.52 \pm 5.12$
& $\mathbf{26.20} \pm 0.03$ & $0.974 \pm 0.000$ & $0.056 \pm 0.000$ & $10.82 \pm 0.29$ & $59.39 \pm 0.63$ \\

\cmark & \cmark & \cmark & \xmark & \cmark & \xmark
& $27.19 \pm 0.06$ & $0.976 \pm 0.000$ & $0.050 \pm 0.001$ & $11.37 \pm 2.09$ & $61.64 \pm 0.51$
& $25.88 \pm 0.04$ & $0.971 \pm 0.000$ & $\mathbf{0.055} \pm 0.001$ & $\mathbf{10.37} \pm 0.62$ & $57.72 \pm 1.14$ \\

\xmark & \cmark & \cmark & \cmark & \cmark & \xmark
& $\mathbf{27.35} \pm 0.79$ & $\mathbf{0.979} \pm 0.001$ & $0.056 \pm 0.019$ & $9.67 \pm 0.05$ & $\mathbf{66.67} \pm 0.13$
& $\mathbf{26.35} \pm 0.22$ & $\mathbf{0.976} \pm 0.000$ & $\mathbf{0.054} \pm 0.001$ & $10.40 \pm 0.46$ & $56.25 \pm 1.05$ \\

\cmark & \xmark & \cmark & \cmark & \cmark & \xmark
& $\mathbf{27.23} \pm 0.06$ & $0.976 \pm 0.000$ & $0.049 \pm 0.000$ & $9.85 \pm 0.87$ & $62.67 \pm 0.72$
& $25.85 \pm 0.04$ & $0.970 \pm 0.000$ & $\mathbf{0.054} \pm 0.001$ & $\mathbf{9.45} \pm 0.12$ & $\mathbf{61.53} \pm 0.60$ \\

\cmark & \cmark & \cmark & \cmark & \cmark & \xmark
& $\mathbf{27.23} \pm 0.14$ & $0.976 \pm 0.001$ & $0.049 \pm 0.001$ & $9.74 \pm 0.41$ & $\mathbf{63.54} \pm 2.77$
& $25.81 \pm 0.02$ & $0.971 \pm 0.000$ & $0.055 \pm 0.000$ & $9.98 \pm 0.26$ & $\mathbf{60.82} \pm 0.10$ \\

\cmark & \cmark & \cmark & \cmark & \cmark & \cmark
& $\mathbf{27.31} \pm 0.16$ & $0.976 \pm 0.001$ & $0.047 \pm 0.002$ & $9.08 \pm 0.53$ & $\mathbf{65.21} \pm 2.15$
& $25.92 \pm 0.09$ & $0.971 \pm 0.000$ & $\mathbf{0.054} \pm 0.000$ & $\mathbf{9.65} \pm 0.27$ & $\mathbf{61.17} \pm 0.89$ \\

\bottomrule
\end{tabular}
}
\end{table*}

\subsection{Comparison with Published Work}

Whitney et al.~\cite{whitney2024learning} is the closest work to ours, as it also learns multi-object collision dynamics using rendering supervision. However, since their code is not publicly available, we are unable to perform a direct quantitative comparison. Instead, we provide a qualitative assessment based on the videos available on their project webpage and our own rollout results. Their demonstrations on the Kubric MOVi-A/B/C datasets~\cite{greff2021kubric} include complex collision events involving multiple objects, where the model appears to struggle with reasoning about multi-object interactions. In contrast, our rollouts exhibit more physically plausible behavior in comparable multi-object scenarios, even though our model is trained on noisy real-world observations. Our rollout videos are available on our project webpage, which is linked in the abstract.

\begin{table*}[t]
\centering
\caption{Performance Comparison on Bowling and Falling Cube Stacks with Published Baseline Reimplemented using our Pipeline and Data Preprocessing Approach}
\scalebox{0.72}{
\begin{tabular}{lc@{\hspace{5pt}}c@{\hspace{5pt}}c@{\hspace{5pt}}c@{\hspace{5pt}}c@{\hspace{5pt}}c@{\hspace{5pt}}c@{\hspace{5pt}}c@{\hspace{5pt}}c@{\hspace{5pt}}c}
\toprule
& \multicolumn{5}{c}{Bowling} & \multicolumn{5}{c}{Falling Cube Stacks} \\
\cmidrule(lr){2-6} \cmidrule(lr){7-11}
Method & PSNR ↑ & SSIM ↑ & LPIPS ↓ & CD ↓ & $\delta_{avg}$ ↑ & PSNR ↑ & SSIM ↑ & LPIPS ↓ & CD ↓ & $\delta_{avg}$ ↑  \\
\midrule
GS-Dynamics* 
& $27.34 \pm 0.06$ & $0.977 \pm 0.000$ & $0.047 \pm 0.001$ & $9.48 \pm 0.13$ & $62.94 \pm 1.56$ 
& $25.85 \pm 0.06$ & $0.971 \pm 0.000$ & $0.054 \pm 0.001$ & $10.07 \pm 0.48$ & $60.83 \pm 0.82$ \\
Ours 
& $27.31 \pm 0.16$ & $0.976 \pm 0.001$ & $0.047 \pm 0.002$ & $9.08 \pm 0.53$ & $65.21 \pm 2.15$ 
& $25.92 \pm 0.09$ & $0.971 \pm 0.000$ & $0.054 \pm 0.000$ & $9.65 \pm 0.27$ & $61.17 \pm 0.89$ \\
\bottomrule
\end{tabular}
}
\label{tbl:dpi}
\vskip -0.1in
\end{table*}

As discussed in Sec.~\ref{sec:related_work}, several prior works~\cite{zhang2024dynamic, lu2025gaussianworldmodel} in robotics train GS-based dynamics models directly from real-world videos. However, these approaches primarily focus on modeling the effects of actions on a single manipulated object. For example, Lu et al.~\cite{lu2025gaussianworldmodel} model interactions using cross-attention between a robot-action token and scene tokens. If the action token is removed, the model no longer captures interactions, making it unsuitable for our multi-object, action-free setting. Similarly, the public implementation of~\cite{zhang2024dynamic} does not support multi-object interactions. Nevertheless, the underlying particle-dynamics module used in~\cite{zhang2024dynamic} is based on DPI~\cite{li2019learning}, which has a publicly available multi-object implementation widely used in intuitive physics benchmarks within simulation environments~\cite{physion_pplus, physion}. Therefore, we reimplement~\cite{zhang2024dynamic} using the DPI formulation and the Gaussian densification scheme proposed in~\cite{zhang2024dynamic}, adapting it to our multi-object, action-free setting. We refer to this variant as GS-Dynamics*. More details on our adaptation of~\cite{zhang2024dynamic, li2019learning} are provided in the supplementary material.

We exclude approaches that use Gaussians generated by GS as input to an MPM-based simulation framework~\cite{Xie_2024_CVPR} from our comparisons. Although these methods can also produce rollouts from input Gaussians, they require per-object physical parameters (e.g., density, Young’s modulus, friction coefficients), which must either be manually specified or estimated via system identification. The performance of MPM is highly sensitive to these parameters. As a result, a fair comparison would require jointly estimating these physical properties from observations using rendering supervision, which remains challenging and has only been explored in simpler settings than our collision scenarios in prior work.

Table~\ref{tbl:dpi} compares our approach with GS-Dynamics*. 
The results indicate that our proposed pipeline also enables training 
GS-Dynamics*
to achieve competitive performance. Our model has some modest numerical advantage to it, while statistical significance is not established.

\section{Limitations} \label{sec:limitations}
A key limitation is the lack of reliable metrics for assessing physical correctness in Gaussian-based dynamics models. In our experiments, rendering-based metrics such as PSNR, SSIM, and LPIPS are not sufficiently discriminative. All methods operate on the same input Gaussians, resulting in visually similar renderings, and the foreground Gaussians occupy only a small portion of the image, further reducing sensitivity. Additionally, when predicted Gaussians do not align with ground-truth object pixels, rendering losses penalize such discrepancies uniformly, without considering whether the predicted configuration is physically plausible. Consequently, these metrics often fail to distinguish between physically meaningful and implausible dynamics.

Geometric metrics such as CD and $\delta_{avg}$ are more sensitive, but we observe that improvements in these metrics do not always correspond to qualitatively more realistic or physically plausible behavior in rollouts. Ideally, evaluation should reflect intuitive physical outcomes, such as whether objects remain stable, how many objects topple, or whether collisions are resolved correctly. However, defining such metrics is challenging in our setting because objects are represented as collections of Gaussians rather than explicit surfaces. As a result, extracting discrete object-level states (e.g., counting the number of cubes remaining on a table or pins that have fallen) is non-trivial and often unreliable.

Developing evaluation metrics that better capture physically meaningful behavior for Gaussian-based representations remains an open problem. We believe this is an important direction for future work, as improved metrics would enable more accurate assessment and comparison of learned dynamics models.

\section{Conclusion}

In this work, we presented the first framework for learning multi-object, action-free 3D collision dynamics from real-world videos using 3D Gaussian trajectories as an intermediate representation. Unlike existing approaches that rely on clean simulation labels, our method learns object interactions purely through rendering supervision and noisy position supervision derived from long-term point tracking and multi-view segmentation. To enable this, we introduced a real-world dataset consisting of around 500 multi-view videos capturing complex collision scenarios, including falling cube stacks and tabletop bowling. Looking forward, our approach opens up new possibilities for learning physically grounded 3D world models from raw and unstructured real-world video observations.
\subsubsection*{Acknowledgements}
This work was partially supported by NSF grants 1751402 and 2321851,
the DARPA TIAMAT grant HR0011-24-9-0423,
and a 2023 Specialty Crop Block Grant from the Oregon Department of Agriculture.
{
    \small
    \bibliographystyle{ieeenat_fullname}
    \bibliography{main}
}

\clearpage
\setcounter{page}{1}
\setcounter{section}{0}
\renewcommand{\thesection}{\Alph{section}}
\maketitlesupplementary

\section{Network Architecture Details}

We adopt a U-Net architecture following \cite{kim2024object}. Each stage of the U-Net contains an interaction block composed of an object PointConv layer followed by a relational PointConv layer. The object PointConv layer may change the spatial resolution depending on the stage via grid downsampling or upsampling, whereas the relational PointConv layer always preserves the input resolution. During downsampling, the feature dimensionality is doubled, and during upsampling it is halved. The specific layer configuration used in this work is provided in Table~\ref{table:unet_architecture}.

\begin{table}[h]
\centering
\caption{U-Net Architecture. The relational PointConv layer in the final decoder block is disabled. We choose $5\,\mathrm{cm}$ as the coarsest resolution, based on the physical size of the real-world objects used in our experiments (approximately $5\!-\!12\,\mathrm{cm}$ along their maximum axis).}
\label{table:unet_architecture}
\resizebox{0.48\textwidth}{!}{
\begin{tabular}{lccc}
\toprule
\textbf{Stage} & \textbf{Interaction Block} & \textbf{Resolution Change} & \textbf{Feature Scaling} \\
\midrule
Input (Pointwise MLP) & -- & $\text{dense} \rightarrow \text{dense}$ & $8 \rightarrow 32$ \\
\midrule
Encoder 
& Block 1 & $\text{dense} \rightarrow \text{dense}$ & $32 \rightarrow 32$ \\
& Block 2 & $\text{dense} \rightarrow 2\,\text{cm}$ & $32 \rightarrow 64$ \\
& Block 3 & $2\,\text{cm} \rightarrow 5\,\text{cm}$ & $64 \rightarrow 128$ \\
\midrule
Bottleneck 
& Block 4 & $5\,\text{cm} \rightarrow 5\,\text{cm}$ & $128 \rightarrow 128$ \\
& Block 5 & $5\,\text{cm} \rightarrow 5\,\text{cm}$ & $128 \rightarrow 128$ \\
\midrule
Decoder 
& Block 6 & $5\,\text{cm} \rightarrow 2\,\text{cm}$ & $128 \rightarrow 64$ \\
& Block 7 & $2\,\text{cm} \rightarrow \text{dense}$ & $64 \rightarrow 32$ \\
& Block 8 & $\text{dense} \rightarrow \text{dense}$ & $32 \rightarrow 32$ \\
\midrule
Prediction Head (Pointwise MLP) & -- & $\text{dense} \rightarrow \text{dense}$ & $32 \rightarrow 3$ \\
\bottomrule
\end{tabular}
}
\end{table}

\section{Training/Test Splits}

As described in the main paper, our dataset contains $210$ falling cube stack sequences and $292$ bowling sequences. We reserve approximately 10\% of the data for testing, resulting in $20$ cube stack test sequences and $30$ bowling test sequences. This leaves $190$ cube stack scenes and $262$ bowling scenes for training, which users may further divide into training and validation sets as needed. All performance numbers reported in the main paper and this supplementary document are evaluated on the test split of each scenario. The exact list of scene names for each split will be made available on our project webpage.

\section{Additional Details on Data Pre-processing}

We feed only the Gaussians corresponding to foreground objects into the network. To achieve this, when running Gaussian Splatting (GS) on each frame to generate input Gaussians, we first apply video object segmentation masks to the input RGB images. This ensures that the GS optimization is performed on images containing only the foreground objects of interest. GS may also generate noisy Gaussians along camera rays (i.e., floaters). We filter these out by requiring each Gaussian to be visible from at least three camera views during GS optimization. Additionally, although the background is masked out, GS may still produce Gaussians in background regions. We remove these by checking their rendering contributions within the segmentation mask.

\section{Additional Details on Integrating an External Baseline into Our Pipeline}
We initially attempted to run GS-Dynamics* on the dense Gaussians produced by Gaussian Splatting \cite{kerbl3Dgaussians} for our dataset. However, the model could not be applied directly to these dense inputs, as it does not scale well to high point densities. This limitation likely stems from its use of an $\varepsilon$-ball neighborhood to define per-point connectivity. With dense Gaussian inputs, each point acquires a significantly larger set of neighbors, leading to substantial computational overhead.

To address this issue, \cite{zhang2024dynamic} select a sparse subset of Gaussians via farthest point sampling and later interpolate the model's predictions back to the original resolution. Following a similar strategy, we apply grid-based downsampling to the original Gaussians—using a grid size of $2\, \mathrm{cm}$ for the cube stack scenario and $4\, \mathrm{cm}$ for the bowling scenario—before feeding them to GS-Dynamics*. We then later apply the Gaussian densification scheme proposed in \cite{zhang2024dynamic} to each predicted object to recover dense Gaussians. This allows us to evaluate the results using rendering-based metrics, as presented in the main paper.

\begin{table*}[t]
\centering
\caption{Ablation results comparing different 4D Gaussian generation methods.} \label{table:ablation_4d}
\scalebox{0.8}{
\begin{tabular}{lcccccccc}
\toprule
& \multicolumn{4}{c}{Bowling} & \multicolumn{4}{c}{Falling Cube Stacks} \\
\cmidrule(lr){2-5} \cmidrule(lr){6-9}
Method & PSNR ↑ & SSIM ↑ & LPIPS ↓ & CD ↓ & PSNR ↑ & SSIM ↑ & LPIPS ↓ & CD ↓ \\
\midrule
Dynamic 3DGS & $25.27 \pm 0.01$ & $0.975 \pm 0.000$ & $0.060 \pm 0.000$ & $15.72 \pm 0.63$ 
        & $24.21 \pm 0.05$ & $0.970 \pm 0.000$ & $0.062 \pm 0.001$ & $13.71 \pm 0.41$ \\

4D GS & $26.98 \pm 0.01$ & $0.980 \pm 0.000$ & $0.047 \pm 0.000$ & $10.86 \pm 0.01$ 
      & $25.27 \pm 0.10$ & $0.973 \pm 0.000$ & $0.057 \pm 0.002$ & $13.12 \pm 1.16$ \\

Ours & $27.57 \pm 0.12$ & $0.979 \pm 0.001$ & $0.044 \pm 0.001$ & $13.98 \pm 2.05$ 
     & $25.83 \pm 0.07$ & $0.971 \pm 0.001$ & $0.055 \pm 0.001$ & $10.05 \pm 0.62$ \\
\bottomrule
\end{tabular}
}
\end{table*}

\section{Ablation on Different 4D Gaussian Generation Methods}
We compare our approach with dynamic-scene GS methods \cite{Wu_2024_CVPR, luiten2023dynamic}, which can produce 3D Gaussian trajectories for both model input and supervision. Because these methods rely on test-time optimization and are slow to run, we construct a \textit{train-mini} set by selecting 70 scenes from each scenario. We apply the GS baselines to the train-mini set and to the test set (20 falling cube stacks and 30 bowling sequences), train our model on train-mini, and evaluate on the corresponding test split.
Each scene is divided into 6-frame clips, and the GS baselines are run on each clip to obtain 6-frame Gaussian trajectories. The first three frames are used as model input and the remaining frames as supervision. Since only short clips are available, long-term tracks are not produced, and we therefore do not use the long-term position metric $\delta_{avg}$ here and only report rendering metrics and Chamfer Distance (CD) here.

The ablation results are presented in Table~\ref{table:ablation_4d}. We observe a consistent performance gap between Dynamic 3D GS and our data across both rendering metrics and Chamfer Distance (CD), with our model outperforming the same architecture trained using Dynamic 3D GS. When comparing 4D GS with our data, the performance gap is smaller. However, we find that training with 4D GS in this setting is generally unstable. In practice, we encountered frequent training failures caused by corrupted or degenerate Gaussian reconstructions, and had to manually remove problematic scenes to complete training. Overall, while 4D GS produces visually high-quality renderings, the underlying 3D Gaussian motion is not always physically consistent. In particular, we observe that 4D GS may generate multiple sets of Gaussians for a fast-moving object, and represent it using different Gaussians across frames. This leads to temporally inconsistent trajectories, which can negatively affect dynamics learning.

\end{document}